\pdfoutput=1

\documentclass[11pt]{article}

\usepackage{acl}

\usepackage{times}
\usepackage{latexsym}
\usepackage{graphicx} 
\usepackage{amsmath}
\usepackage{amsfonts}
\usepackage{bm}
\usepackage[linesnumbered,ruled,vlined]{algorithm2e}
\usepackage{comment}

\usepackage[T1]{fontenc}

\usepackage[utf8]{inputenc}

\usepackage{microtype}

\title{Creative Painting with Latent Diffusion Models}

\author{Xianchao Wu \\
NVIDIA \\
  \texttt{xianchaow@nvidia.com, wuxianchao@gmail.com} }

\begin{document}
\maketitle
\begin{abstract}
Artistic painting has achieved significant progress during recent years. Using an autoencoder to connect the original images with compressed latent spaces and a cross attention enhanced U-Net as the backbone of diffusion, latent diffusion models (LDMs) have achieved stable and high fertility image generation. In this paper, we focus on enhancing the creative painting ability of current LDMs in two directions, textual condition extension and model retraining with Wikiart dataset. Through textual condition extension, users' input prompts are expanded with rich contextual knowledge for deeper understanding and explaining the prompts. Wikiart dataset contains 80K famous artworks drawn during recent 400 years by more than 1,000 famous artists in rich styles and genres. Through the retraining, we are able to ask these artists to draw novel and creative painting on modern topics. Direct comparisons with the original model show that the creativity and artistry are enriched.

\end{abstract}

\section{Introduction}

Artistic painting has achieved significant progress during recent years thanks to the appearing of hundreds of GAN variants \cite{gan_survey1_DBLP:journals/corr/abs-2006-05132, gan_survey2}. However, adversarial training has been reported to be notoriously unstable and can lead to mode collapse. To escape from adversarial training and inspired by non-equilibrium thermodynamics, diffusion probabilistic models \cite{DPM2015_DBLP:journals/corr/Sohl-DicksteinW15}, such as noise-conditional score network (NCSN) \cite{ScoreMatching_NEURIPS2019_3001ef25}, denoising diffusion probabilistic models (DDPM) \cite{DDPM_DBLP:journals/corr/abs-2006-11239}, stable diffusion models in latent spaces \cite{stablediffusion_DBLP:journals/corr/abs-2112-10752} have achieved GAN-level sample quality without adversarial training. These diffusion models are appealing with rather flexible model architectures, exact log-likelihood computation, and inverse problem solving without re-training models. 

There are two Markov chain style processes in a typical diffusion model. The first process is a \emph{forward diffusion process} which appends multiple-scale random noise to a given data sample ``step by step'' or ``in jump'' until the disturbed sample slip into a predefined isotropic Gaussian distribution. This process does not include trainable parameters. The second process is a \emph{reverse diffusion process} which generates a target distribution data sample from pure noise guided by some (user-input) pre-given conditions. A parameterized deep learning model is required in this reverse process. 

Intuitively speaking, the forward diffusion process can be recognized as ``directional blasting of a building'' $\textbf{x}_0$ to ``ruins with dusts'' $\textbf{x}_T$. The learning algorithm is a \emph{reverse engineering} which learns how to (re-)construct a building (expressed by $p_\theta(\textbf{x}_{t-1}|\textbf{x}_t)$ with a parameter set $\theta$ and $t\in\{1, ..., T\}$) from each step of \emph{inverse} directional blasting (expressed by $q(\textbf{x}_{t-1}|\textbf{x}_{t}, \textbf{x}_0)$) of each given building sample $\textbf{x}_0$. In one step of this reverse engineering, $\textbf{x}_{t-1}$ represents ``one complete wall'' in a building and $\textbf{x}_t$ represents ``concrete and sands'' that can be used to construct the complete wall $\textbf{x}_{t-1}$ in a reconstruction process or can be obtained from the complete wall $\textbf{x}_{t-1}$ in a forward ``blasting'' process. The reconstruction process is learned from the blasting process with targets such as noise prediction in DDPM \cite{DDPM_DBLP:journals/corr/abs-2006-11239} or score prediction using score matching strategy in NCSN \cite{ScoreMatching_NEURIPS2019_3001ef25}.

We follow a recent impressive work of high-resolution image synthesis with LDMs by given textual or visual conditions\footnote{\url{https://github.com/CompVis/stable-diffusion}} \cite{stablediffusion_DBLP:journals/corr/abs-2112-10752}. There are several proposals in this LDM. The first proposal is applying the encoder part of a pretrained autoencoder to project images into low-dimension latent spaces and then perform diffusion/construction processes. Training diffusion models on such a low-dimension representation space allows us to reach a near-optimal point between computation complexity reduction and detail preservation to boost virtual fidelity of constructed images. The second is a cross-attention-enhanced \cite{transformer_NIPS2017_3f5ee243} U-Net framework \cite{Unet_DBLP:journals/corr/RonnebergerFB15} in the diffusion model where general conditioning inputs such as text or bounding boxes are taken as \emph{memory} (i.e., keys and values in the cross-attention layers) for the query (latent representations of images to be generated) to retrieve information on. Finally, the decoder module in the autoencoder is applied to recover the target image into high-resolution.

We aim at improving the \emph{creativity} of image synthesis, or painting, using conditional LDMs. It is relatively difficult to precisely define the concept of creativity since it is  subjective and influenced by culture, history, and region. The color, style, objects included in painting reflect rich emotions of numerous topics. For example, when we are given a textual condition, ``a painting of a virus monster playing guitar'', we can recognize noun entities such as ``virus monster'' and ``guitar'' and a verbal action ``playing''. What are the emotions involved in this textual hint? Happy, surprise and funny should be the major emotions. The painting requires less imaginations since we should better include the entries with a determined action. 

However, there are challenges for the models to draw painting for rather high-level topics such as ``urbanization of China'' or ``Asian morning''. These textual hints should be enriched and extended with concrete objects and actions to tell a story in a painting or in a series of paintings. Extensions to ``urbanization of China'' include ``originally a collection of fishing villages, Shenzhen rapidly grew to be one of the largest cities in China'', ``a train runs on the snow-capped mountains of the Qinghai-Tibet Plateau'', and ``left-behind children running in wheat-field''. Given an initial textual hint, we leverages Wikipedia and large-scale pretrained language models to execute this extension.

In addition, we retrain existing checkpoints by the WikiArt paintings dataset\footnote{\url{https://www.wikiart.org/} and can be downloaded from \url{https://archive.org/download/wikiart-dataset/wikiart.tar.gz}} which has a collection of 81,444 fine-art paintings from 1,119 artists, ranging from fifteenth century to modern times. This dataset contains 27 different styles (e.g., \emph{Minimalism}, \emph{Symbolism}, \emph{Realism}) and 45 different genres. As far as our knowledge, it is currently the largest digital art datasets publicly available for research usage. This dataset was used to train an ArtGAN \cite{artgan_DBLP:journals/corr/TanCAT17} where conditions such as categorical label information was used for artwork synthesis. In this paper, we embed the textual information of artists, year, styles, and genres as additional conditions to the LDM. Through this way, we can determine explicitly to invite Vincent van Gogh or Rembrant to help us drawing artworks of modern topics such as ``urbanization of China''. 

This paper is organized as follows. In Section \ref{sec:background}, we briefly review the background knowledge required for understanding the stable diffusion models \cite{stablediffusion_DBLP:journals/corr/abs-2112-10752}. In particular, we describe the two processes defined in DDPM \cite{DDPM_DBLP:journals/corr/abs-2006-11239}, the autoencoder framework and loss functions used in it \cite{taming_DBLP:journals/corr/abs-2012-09841}, cross attention enhanced U-Net which acts as the backbone of the diffusion model, and pseudo numerical methods integrated with DDIMs for fast sampling. In Section \ref{sec:text_extend}, we describe our proposal of extending users' prompts by pretrained language models and existing knowledge resources. In Section \ref{sec:retrain_wikiart}, we show detailed information of the Wikiart dataset and our pipeline of retraining. We describe the experiments in Section \ref{sec:experiments} and finally conclude in Section \ref{sec:conclusion}.

\section{Background}\label{sec:background}

Diffusion models have been successfully used in image generation \cite{stablediffusion_DBLP:journals/corr/abs-2112-10752}, text-to-speech synthesis \cite{grad_tts_DBLP:journals/corr/abs-2105-06337, diff_tts_https://doi.org/10.48550/arxiv.2104.01409}, sing synthesis and conversion \cite{sing_conv_diff_https://doi.org/10.48550/arxiv.2105.13871, learn2sing_https://doi.org/10.48550/arxiv.2203.16408}, music generation \cite{music_diffusion_DBLP:journals/corr/abs-2103-16091} and healthcare Medical Anomaly Detection \cite{medical_diff_https://doi.org/10.48550/arxiv.2203.04306}. Surveys can be find in \cite{survey_vision_diffusion_https://doi.org/10.48550/arxiv.2209.04747, survey_generative_diffusion_https://doi.org/10.48550/arxiv.2209.02646, diff_beida_survey_https://doi.org/10.48550/arxiv.2209.00796}.

We limit our discussion to text-to-image generation by leveraging the LDMs \cite{stablediffusion_DBLP:journals/corr/abs-2112-10752} and existing checkpoints\footnote{\url{https://huggingface.co/CompVis/stable-diffusion-v-1-4-original}}.
We briefly review the core processes and target objectives of DDPMs \cite{DDPM_DBLP:journals/corr/abs-2006-11239} that are used in LDMs. In addition, autoencoders enhanced with KL-divergence, cross-attention embedded U-Net \cite{Unet_DBLP:journals/corr/RonnebergerFB15,transformer_NIPS2017_3f5ee243}, CLIP pretrained language models \cite{clip_DBLP:journals/corr/abs-2103-00020} and sampling algorithms such as that used in denoising diffusion implicit models (DDIMs) \cite{ddim_DBLP:journals/corr/abs-2010-02502} and pseudo numerical methods \cite{pseudo_https://doi.org/10.48550/arxiv.2202.09778} will be briefly reviewed.

\subsection{DDPM}

\begin{figure}[t]
  \centering
  \includegraphics[width=7.5cm]{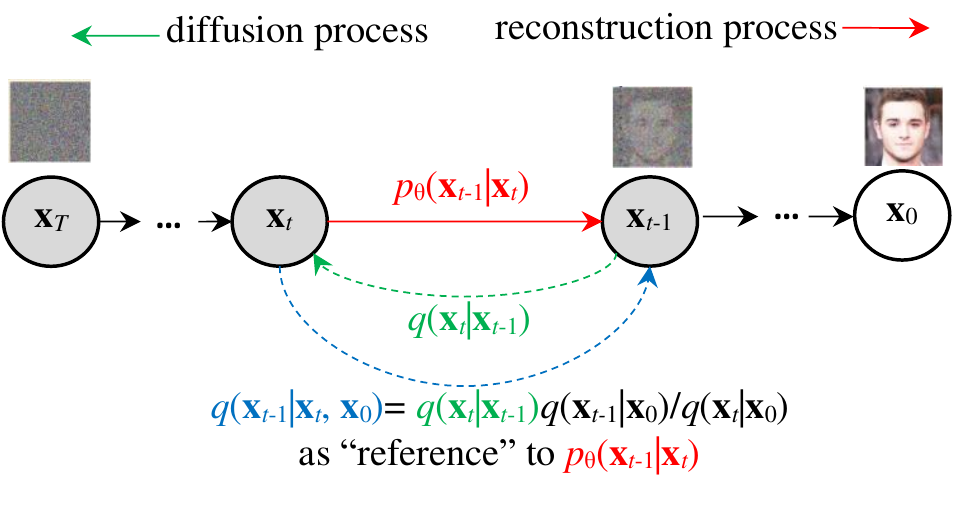}
  \caption{The Markov chain of forward diffusion (backward reconstruction) process of generating a sample by step-by-step adding (removing) noise. Image adapted from \cite{DDPM_DBLP:journals/corr/abs-2006-11239}.}
  \label{fig:ddpm_two_processes}
\end{figure}

Given a data point $\textbf{x}_0$ sampled from a real data distribution $q(\textbf{x})$ ($\textbf{x}_0 \sim q(\textbf{x})$), \newcite{DDPM_DBLP:journals/corr/abs-2006-11239} define a \emph{forward diffusion process} in which small amount of Gaussian noise is added to sample $\textbf{x}_0$ in $T$ steps to obtain a sequence of noisy samples $\textbf{x}_0, ..., \textbf{x}_T$. A predefined (hyper-parameter) variance schedule $\{ \beta_t \in (0, 1) \}_{t=1}^T$ controls the step sizes:
\begin{align}
    q(\textbf{x}_t | \textbf{x}_{t-1}) & = \mathcal{N}(\textbf{x}_t; \sqrt{1-\beta_t}\textbf{x}_{t-1}, \beta_t \textbf{I}); \\
    q(\textbf{x}_{1:T} | \textbf{x}_0) & := \prod_{t=1}^Tq(\textbf{x}_t | \textbf{x}_{t-1}). \label{eq:q_distribution}
\end{align}
When $T \rightarrow \infty$, $\textbf{x}_T$ is equivalent to following an isotropic Gaussian distribution. Note that, there are no trainable parameters used in this forward diffusion process.

Let $\alpha_t = 1 - \beta_t$ and $\bar{\alpha}_t = \prod_{i=1}^t \alpha_i$, we can express an arbitrary step $t$'s diffused sample $\textbf{x}_t$ by the initial data sample $\textbf{x}_0$:
\begin{equation}
    \textbf{x}_t = \sqrt{\bar{\alpha}_t} \textbf{x}_0 + \sqrt{1 - \bar{\alpha}_t} \bm{\epsilon}_t. \label{eq:xt_x0_relation}
\end{equation}
Here, noise $\bm{\epsilon}_t \sim \mathcal{N}(0, \textbf{I})$ shares the same shape with $\textbf{x}_0$ and $\textbf{x}_t$.

In order to reconstruct from a Gaussian noise input $\textbf{x}_T \sim \mathcal{N}(0, \textbf{I})$, we need to learn a model $p_\theta$ to approximate the conditional probabilities to run the \emph{reverse diffusion process}:
\begin{align}
    p_\theta(\textbf{x}_{t-1} | \textbf{x}_{t}) & = \mathcal{N}(\textbf{x}_{t-1}; \bm{\mu}_\theta(\textbf{x}_t, t), \bm{\Sigma}_\theta(\textbf{x}_t, t)); \\
    p_\theta(\textbf{x}_{0:T}) & := p(\textbf{x}_T)\prod_{t=1}^Tp_\theta(\textbf{x}_{t-1} | \textbf{x}_{t}). \label{eq:p_theta_distribution}
\end{align}

Note that the reverse conditional probability is tractable by first applying Bayes' rule to three Gaussian distributions and then completing the ``quadratic component'' in the $\text{exp}(\cdot)$ function:
\begin{align}
    q(\textbf{x}_{t-1} | \textbf{x}_{t}, \textbf{x}_0) & = \mathcal{N}(\textbf{x}_{t-1}; \tilde{\bm{\mu}}_t(\textbf{x}_t, \textbf{x}_0), \tilde{\beta}_t\textbf{I}) \\
    & = q(\textbf{x}_{t} | \textbf{x}_{t-1}, \textbf{x}_0)\frac{q(\textbf{x}_{t-1} | \textbf{x}_0)}{q(\textbf{x}_{t} | \textbf{x}_0)} \\
    & \propto \text{exp}(-\frac{1}{2\tilde{\beta}_t}(\textbf{x}_{t-1} - \tilde{\bm \mu}_t)^2).
\end{align}
Here, variance $\tilde{\beta}_t$ is a scalar and mean $\tilde{\bm \mu}_t$ depends on $\textbf{x}_t$ and noise $\bm{\epsilon}_t$:
\begin{align}
    \tilde{\beta}_t & = \frac{1-\bar{\alpha}_{t-1}}{1-\bar{\alpha}_{t}}\beta_t; \\
    \tilde{\bm \mu}_t & = \frac{1}{\sqrt{\alpha_t}}(\textbf{x}_t - \frac{1-{\alpha}_{t}}{\sqrt{1-\bar{\alpha}_{t}}}\bm{\epsilon}_t).
\end{align}
Intuitively, $q(\textbf{x}_{t-1} | \textbf{x}_{t}, \textbf{x}_0)$ acts as a \emph{reference} to learn $p_\theta(\textbf{x}_{t-1} | \textbf{x}_{t})$. We can use the variational lower bound (VLB) to optimize the negative log-likelihood:
\begin{multline}
    -\text{log}p_\theta(\textbf{x}_0) \leq -\text{log}p_{\theta}(\textbf{x}_0) + \\ D_{\text{KL}}(q(\textbf{x}_{1:T}|\textbf{x}_0) \parallel p_\theta(\textbf{x}_{1:T}|\textbf{x}_0)).
\end{multline}

Using the definitions of $q(\textbf{x}_{1:T}|\textbf{x}_0)$ in Equation \ref{eq:q_distribution} and $p_\theta(\textbf{x}_{0:T})$ in Equation \ref{eq:p_theta_distribution}, a loss item $L_t$ ($1 \leq t \leq T-1$) is expressed by:
\begin{align}
    L_t & = D_{\text{KL}}(q(\textbf{x}_{t}|\textbf{x}_{t+1}, \textbf{x}_0) \parallel p_\theta(\textbf{x}_{t}|\textbf{x}_{t+1})) \\
    & = \mathbb{E}_{\textbf{x}_0, \bm{\epsilon}_t} \left [ \frac{1}{2\parallel \bm{\Sigma}_\theta(\textbf{x}_t, t)\parallel_2^2} \parallel \tilde{\bm{\mu}}_t - \bm{\mu}_\theta(\textbf{x}_t, t)\parallel^2 \right ]. \nonumber
\end{align}
We further reparameterize the Gaussian noise term instead to predict $\bm{\epsilon}_t$ from time step $t$'s input $\textbf{x}_t$ and use a simplified objective that ignores the weighting term:
\begin{align}
    L_t^{\text{simple}} & = \mathbb{E}_{t \sim [1, T], \textbf{x}_0, \bm{\epsilon}_t} \left [ \parallel \bm{\epsilon}_t - \bm{\epsilon}_\theta(\textbf{x}_t, t)  \parallel^2 \right ] \\
    & = \mathbb{E}\left [ \parallel \bm{\epsilon}_t - \bm{\epsilon}_\theta(\sqrt{\bar{\alpha}_t}\textbf{x}_0+\sqrt{1-\bar{\alpha}_t}\bm{\epsilon}_t, t)  \parallel^2 \right ]. \nonumber
\end{align}

In \cite{stablediffusion_DBLP:journals/corr/abs-2112-10752}, LDMs are proposed so that the diffusion processes are performed in compressed latent spaces through a pretrained autoencoder $\mathcal{E}(\textbf{x}_0)$:
\begin{align}
    L_t^{\text{LDM}} & = \mathbb{E}_{\textbf{z}_0=\mathcal{E}(\textbf{x}_0), \bm{\epsilon}_t, t}\left [ \parallel \bm{\epsilon}_t - \bm{\epsilon}_\theta(\textbf{z}_t, t)  \parallel^2 \right ] \\
    & = \mathbb{E}\left [ \parallel \bm{\epsilon}_t - \bm{\epsilon}_\theta(\sqrt{\bar{\alpha}_t}\textbf{z}_0+\sqrt{1-\bar{\alpha}_t}\bm{\epsilon}_t, t)  \parallel^2 \right ]. \nonumber
\end{align}
In order to perform condition-based image synthesis, a pre-given textual prompt (or other formats such as layout) $y$ is first encoded by a domain specific encoder $\tau_\theta(y)$ and then sent to the model to predict $\bm{\epsilon}_\theta$:
\begin{equation}
    L_t^{\text{LDM}} = \mathbb{E}_{\mathcal{E}(\textbf{x}_0), \bm{\epsilon}_t, t}\left [ \parallel \bm{\epsilon}_t - \bm{\epsilon}_\theta(\textbf{z}_t, t, \tau_\theta(y))  \parallel^2 \right ]. \label{eq:ldm}    
\end{equation}
Here, $\tau_\theta(y)$ acts as memory (key and value) in the cross-attention mechanism \cite{transformer_NIPS2017_3f5ee243} and can be jointly trained together with $\bm{\epsilon}_\theta$'s U-Net framework \cite{Unet_DBLP:journals/corr/RonnebergerFB15} from image-conditioning pairs. In the text-to-image generation task of \cite{stablediffusion_DBLP:journals/corr/abs-2112-10752}, a 12-layer transformer with a hidden dimension of 768 is used\footnote{\url{https://huggingface.co/openai/clip-vit-large-patch14}} \cite{clip_DBLP:journals/corr/abs-2103-00020} to encode textual prompts. 

\subsection{Autoencoder with KL-divergence}

The autoencoder is pretrained \cite{taming_DBLP:journals/corr/abs-2012-09841} beforehand and used directly for encoding the original data sample into latent space and for decoding the reconstructed $\textbf{z}_0$ back to the original sizes of $\textbf{x}_0$. In order to combine the effectiveness of the inductive bias of CNNs with the expressivity of transformers, both the encoder ($\mathcal{E}$) and the decoder (or, generator, $\mathcal{G}$) parts of the autoencoder use resnet blocks and self-attention blocks. 

Adversarial learning is used to train this vector quantised GAN framework with a combination of several losses: 

(1) a reconstruction loss: 
\begin{equation}
    \parallel \textbf{x} - \mathcal{G} (\textbf{q}(\mathcal{E}(\textbf{x})))\parallel^2,
\end{equation}
 where $\textbf{q}(\cdot)$ is element-wise quantization.
 
(2) a KL loss on the diagonal Gaussian distribution constructed from $\textbf{q}(\mathcal{E}(\textbf{x}))$ = [$\bm{\mu}$; $\text{log}\bm{\sigma}^2$]:
 \begin{equation}
     \sum_{c, h, w}(\bm{\mu}^2 + \bm{\sigma}^2 - 1 - \text{log}\bm{\sigma}^2)/2
 \end{equation}
 where $c$ is channel number, $h$ is height and $w$ is width. The output tensor $\textbf{q}(\mathcal{E}(\textbf{x}))$ is separated into two parts (e.g., from (6, 64, 64) to two (3, 64, 64) shape tensors) for the mean and the log of the variance of the Gaussian distribution. 
 
 (3) a GAN loss which includes the following component:
 \begin{equation}
     \text{log}\mathcal{D}(\textbf{x})+\text{log}(1-\mathcal{D}(\mathcal{G} (\textbf{q}(\mathcal{E}(\textbf{x})))). 
 \end{equation}
Here, $\mathcal{D}$ stands for a patch-based discriminator that aims to differentiate between real and reconstructed images. Adaptive weight is used to combine these losses and more details can be found in \cite{taming_DBLP:journals/corr/abs-2012-09841}.

\subsection{U-Net with Cross Attention}

In \cite{stablediffusion_DBLP:journals/corr/abs-2112-10752}, a U-Net with a multi-head cross attention mechanism \cite{transformer_NIPS2017_3f5ee243} is used to predict $\bm{\epsilon}_\theta$ with a MSE loss for training (Equation \ref{eq:ldm}). In a typical U-Net implementation, there are five blocks, a \emph{time embedding block} that embeds an input time step $t$, \emph{input/middle/output blocks} that perform convolutional and self-attention based representations of $\textbf{z}_t$ and their cross attentions with conditional memory $\tau_\theta(y)$, and finally a \emph{out block} that projects the result tensor back to the shape of $\textbf{z}_t$. 

The \emph{input block} performs a down sampling with a stack of ``resnet + spatial transformer'' modules (e.g., 12 modules from (channel, height, width) shape of from (4, 64, 64) to (1280, 8, 8)). Then, the \emph{middle block} with ``resnet + transformer + resnet'' modules links the \emph{input} and \emph{output blocks} without changing the shape of the tensor. Next, the \emph{output block} performs a up sampling with the same number of modules of the input block (e.g., 12 modules from shape (1280, 8, 8) to (320, 64, 64)). There are residual-style shortcut links here: each module's output are sent respectively from the \emph{input block} to the \emph{output block} with same level. The final \emph{out block} uses a 2D convolutional layer to project the hidden channel number (e.g., 320) back to the original channel number (e.g., 4).

\subsection{DDIMs and Pseudo Numerical Methods}

DDIMs \cite{ddim_DBLP:journals/corr/abs-2010-02502} generalizes DDPMs via a class of non-Markovian diffusion processes that lead to the same training objective and give rise to implicit models that generate high quality samples much faster. In the non-Markovian forward process, a real vector $\sigma \in \mathbb{R}_{\geq 0}^T$ is introduced to index a family of \emph{inference} distributions:
\begin{align}
    q_\sigma(\textbf{x}_{1:T} | \textbf{x}_0) & := q_\sigma(\textbf{x}_{T} | \textbf{x}_0) \prod_{t=2}^Tq_\sigma(\textbf{x}_{t-1} | \textbf{x}_{t}, \textbf{x}_{0}); \nonumber \\
    q_\sigma(\textbf{x}_{T} | \textbf{x}_0) & = \mathcal{N}(\sqrt{\bar{\alpha}_T}\textbf{x}_0, (1-\bar{\alpha}_T)\textbf{I}); \nonumber \\
    q_\sigma(\textbf{x}_{t-1} | \textbf{x}_{t}, \textbf{x}_{0}) & = \mathcal{N}(\tilde{\bm{\mu}}(\textbf{x}_0, \textbf{x}_t, \sigma_t), \sigma_t^2\textbf{I}); \nonumber \\
    \tilde{\bm{\mu}}(\textbf{x}_0, \textbf{x}_t, \sigma_t) & = \sqrt{\bar{\alpha}_{t-1}}\textbf{x}_0 +  \nonumber \\
    & \sqrt{1-\bar{\alpha}_{t-1} - \sigma_t^2}\frac{\textbf{x}_t - \sqrt{\bar{\alpha}_t}\textbf{x}_0}{\sqrt{1-\bar{\alpha}_t}}. \nonumber
\end{align}
The mean function $\tilde{\bm{\mu}}(\textbf{x}_0, \textbf{x}_t, \sigma_t)$ is chosen to ensure that $q_\sigma(\textbf{x}_{t} | \textbf{x}_0) = \mathcal{N}(\sqrt{\bar{\alpha}_t}\textbf{x}_0, (1-\bar{\alpha}_t)\textbf{I})$ without depending on $\sigma$ anymore.

In the generative process of DDIM, the \emph{denoised observation} $\textbf{x}_0$ is predicted from pre-given $\textbf{x}_t$ (reverse usage of Equation \ref{eq:xt_x0_relation}):
\begin{equation}
    f_\theta(\textbf{x}_t, t) := (\textbf{x}_t - \sqrt{1-\bar{\alpha}_t}\bm{\epsilon}_\theta(\textbf{x}_t, t))/\sqrt{\bar{\alpha}_t}. \nonumber
\end{equation}
Then, a sample $\textbf{x}_{t-1}$ can be generated from $\textbf{x}_t$ via:
\begin{align}
    \textbf{x}_{t-1} & =\sqrt{\bar{\alpha}_{t-1}}f_\theta(\textbf{x}_t, t)  \nonumber \\
    & + \sqrt{1-\bar{\alpha}_{t-1} - \sigma_t^2}\bm{\epsilon}_\theta(\textbf{x}_t, t) + \sigma_t\bm{\epsilon}_t. 
\end{align}
When $\sigma_t=0$ for all $t$, the coefficient of $\bm{\epsilon}_t$ becomes zero and samples are generated from $\textbf{x}_T$ to $\textbf{x}_0$ with a fixed procedure. The $\text{DDIM}(\cdot)$ is thus defined as:
\begin{equation}
    \textbf{x}_{t-1}, f_\theta(\textbf{x}_t, t) = \text{DDIM}(\textbf{x}_t, \bm{\epsilon}_t,t). \label{eq:ddim_xt_to_xtm1}
\end{equation}

To accelerate the reconstruction process and keep the sample quality, DDIMs (Equation \ref{eq:ddim_xt_to_xtm1}) are included in pseudo numerical methods \cite{pseudo_https://doi.org/10.48550/arxiv.2202.09778} which treat DDPMs as solving differential equations on manifolds. In \cite{stablediffusion_DBLP:journals/corr/abs-2112-10752}'s code implementation\footnote{\url{https://github.com/CompVis/stable-diffusion/blob/main/ldm/models/diffusion/plms.py\#L218-L232}} (Algorithm \ref{alg:pndm_comb_ddim_plms}), a linear multi-step algorithm, the Adams-Moulton method\footnote{\url{https://en.wikipedia.org/wiki/Linear\_multistep\_method\#CITEREFHairerN\%C3\%B8rsettWanner1993}}, is used. This pseudo numerical algorithm includes a gradient part of 2nd order pseudo improved Euler,  2nd/3rd/4th order Adams-Bashforth methods, and a transfer part of DDIM. Here, the discrete indices $t-1, t+1$ stand for next (e.g., from $T$ to $T-1$) and former time steps, respectively.

\begin{algorithm}[t]
\caption{Pseudo linear multi-step (PLMS) algorithm enhanced by DDIM }\label{alg:pndm_comb_ddim_plms}
 $\textbf{x}_T \sim \mathcal{N}(0, \textbf{I})$\;
 \For{$t=T, T-1, ..., 1$}{
  $e_t = \bm{\epsilon}_\theta(\textbf{x}_t, t)$\;
  \uIf{t == T}{
    \# pseudo improved Euler-2nd\;
    $\textbf{x}_{t-1}, f_\theta(\textbf{x}_{t},t) = \text{DDIM}(\textbf{x}_t, e_t, t)$\;
    $e_{t-1} = \bm{\epsilon}_\theta(\textbf{x}_{t-1}, t-1)$\;
    $e'_t = (e_t + e_{t-1})/2$\;
  }
  \uElseIf{t == T-1}{
    \# PLMS-2nd (Adams-Bashforth) \;
    $e'_t = (3e_t - e_{t+1})/2$\;
  }
  \uElseIf{t == T-2}{
    \# PLMS-3rd (Adams-Bashforth) \;
    $e'_t = (23e_t - 16e_{t+1} + 5e_{t+2})/12$\;
  }
  \Else{
    \# PLMS-4th (Adams-Bashforth) \;
    $e'_t = (55e_t - 59e_{t+1} + 37e_{t+2}-9e_{t+3})/24$\;
  }
  $\textbf{x}_{t-1}, f_\theta(\textbf{x}_{t},t) = \text{DDIM}(\textbf{x}_t, e'_t, t)$\;
 }
 return $\textbf{x}_0$\;
\end{algorithm}

\section{Textual Condition Extension}\label{sec:text_extend}

\begin{figure}
  \centering
  \includegraphics[width=7.5cm]{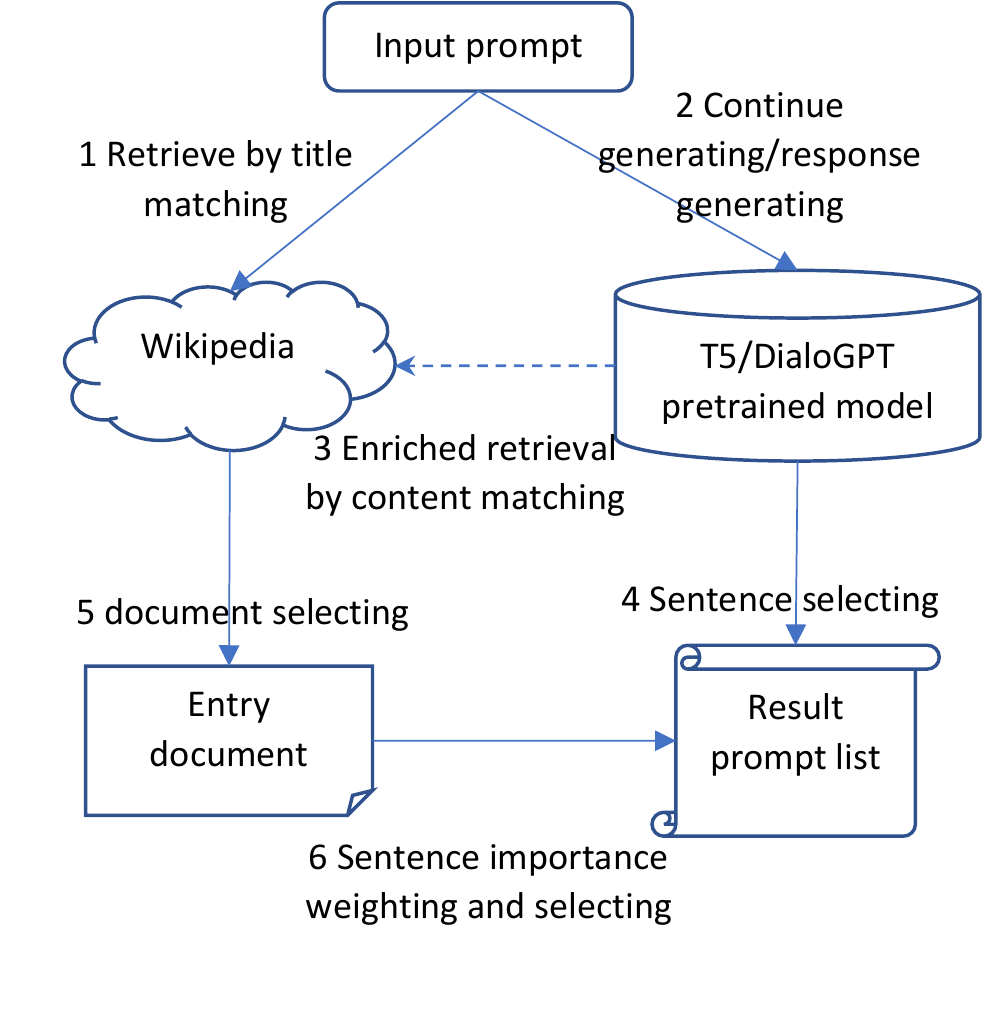}
  \caption{The textual prompt extension pipeline by retrieving wikipedia and continue generating by T5/DialoGPT pretrained language models \cite{t5_DBLP:journals/corr/abs-1910-10683, dialogpt_DBLP:journals/corr/abs-1911-00536}.}
  \label{fig:text_extension_framework}
\end{figure}

We perform textual condition extension by leveraging wikipedia as the knowledge base and large-scale pretrained language models as implicit knowledge graphs. The pipeline is depicted in Figure \ref{fig:text_extension_framework}. Given a textual prompt, we first match it with the title list in wikipedia. At the same time, the input prompt is sent to (1) a pretrained language model, T5 \cite{t5_DBLP:journals/corr/abs-1910-10683}, to continue writing by taking the given prompt as a prefix hint and to (2) a pretrained dialog model, DialoGPT\footnote{\url{https://github.com/microsoft/DialoGPT}} \cite{dialogpt_DBLP:journals/corr/abs-1911-00536} that takes the input prompt as ``query'' and consequently generate ``responses''. 

Wikipedia's titles and contents are used for matching the input prompt and T5/DialoGPT's outputs. We use BM25 \cite{bm25_robertson2009probabilistic} here to simplify the matching process. From the result document(s), we further compute sentence importance to rank their content fertility and the relationship with the initial prompt. We use the (English) text part of LAION-5B\footnote{\url{https://laion.ai/blog/laion-5b/}} and Wikiart to train a TF-IDF model and then use it to score the prompts in the result prompt list. With a higher score, we subjectively believe that the prompt can possibly yield better images. To score the ``relationship'' with the initial prompt $u$, we embed a pair of initial and result prompts by T5 and compute their cosine similarity. Thus, the importance of a result prompt $v$ is computed by:
\begin{equation}
    w(v) = \text{TFIDF}(v) + \lambda_1 \text{Cos}(\text{T5}(u), \text{T5}(v)).
\end{equation}
Here, $\lambda_1$ stands for a hyper-parameter to balance the scale of two scores.

As former mentioned, we encourage the result prompts to include spacial and temporal information. We leverage a named entity recognizer\footnote{\url{https://github.com/kamalkraj/BERT-NER}} and regular expressions to recognize place/region names, addresses, time, and date. The number of spacial and temporal entities discounted by a hyper parameter $\lambda_2$ is added with $w(v)$ for the final scoring of a prompt.

\section{Retraining with WikiArt}\label{sec:retrain_wikiart}

\begin{figure*}[t]
  \centering
  \includegraphics[width=16cm]{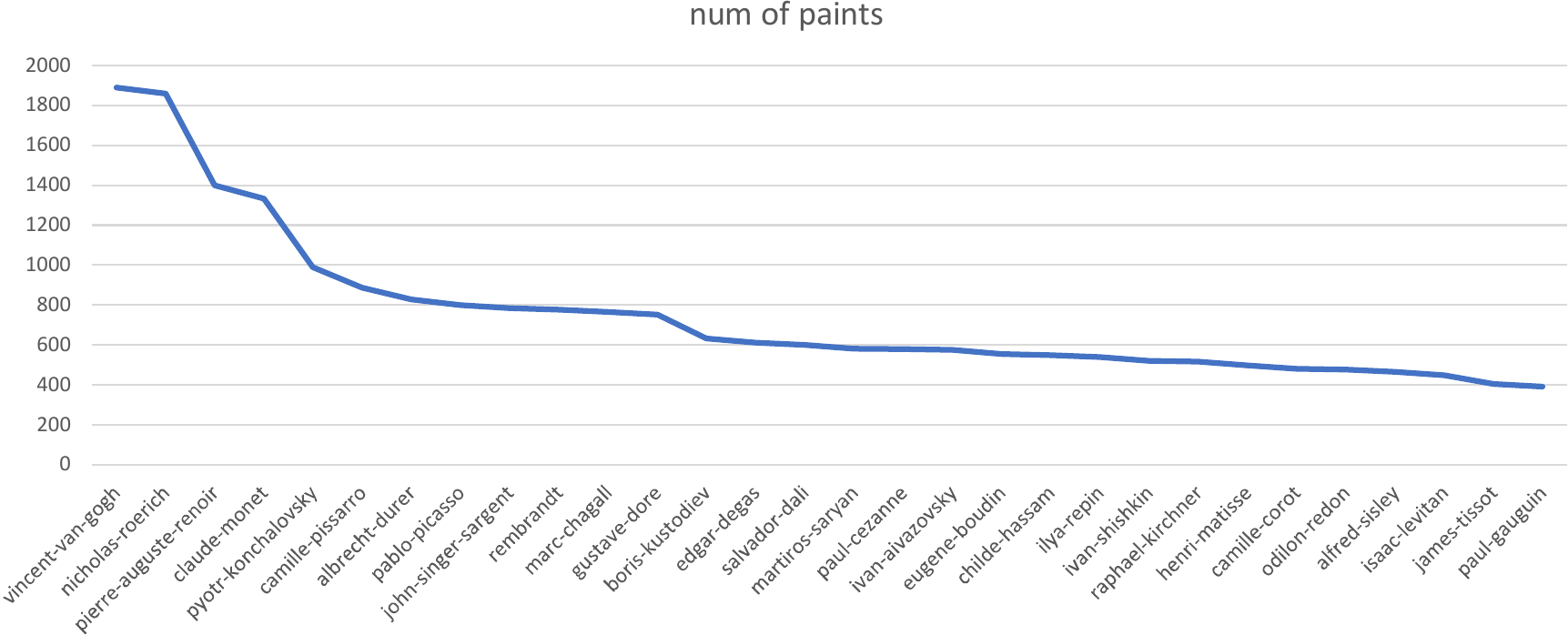}
  \caption{Top-30 artists and their painting numbers in Wikiart.}
  \label{fig:top-30-artists}
\end{figure*}

Different artists have quite different number of paints in WikiArt dataset. The top-3 artists are Vincent van Gogh, Nicholas Roerich, and Pierre Auguste Renoir with 1,889, 1,860, and 1,400 paints, respectively. The top-10, top-20, and top-30 artists share 14.18\%, 21.80\%, and 27.62\% of the samples, respectively. Figure \ref{fig:top-30-artists} shows the distribution of the number of paints and their authors in Wikiart.

We first retrain the CLIP text encoder with the same tokenizer with the LDM fixed. This stage is expected to map the captions used in Wikiart to stable diffusion's latent space. Then, we fine-tune the text encoder and the LDM jointly. This stage is expected to help the LDM to enrich its knowledge of artworks from different artists, in different styles and genres.

\section{Experiments}\label{sec:experiments}

We use a DGX-A100-80GB server with 8 NVIDIA A100-80GB GPU cards. The original code and settings of stable diffusion model's checkpoint v1-4 is reused. During inferencing, single GPUs are used with ddim\_eta=1.0, ddim\_steps=200, height and width are both 512, and scale is set to be 5.0.

\subsection{Direct Comparison with Original LDMs}

\begin{figure*}
  \centering
  \includegraphics[width=16cm]{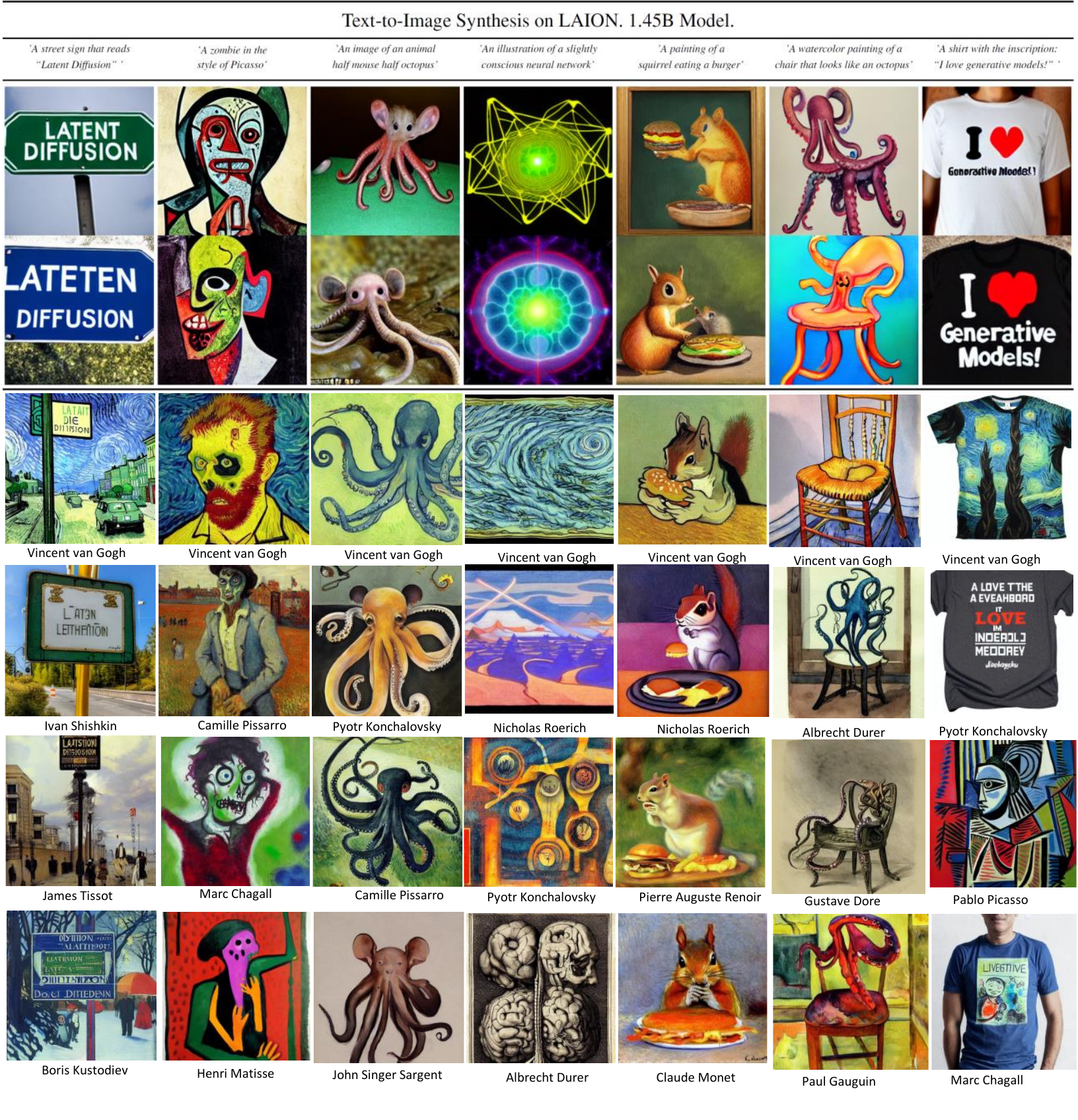}
  \caption{Direct comparison with same prompts used in \cite{stablediffusion_DBLP:journals/corr/abs-2112-10752} yet different artists.}
  \label{fig:compare-top-30-artists}
\end{figure*}

Figure \ref{fig:compare-top-30-artists} directly compares the images generated by the original model and that retrained under Wikiart. We used the same prompts as described in \cite{stablediffusion_DBLP:journals/corr/abs-2112-10752}. For direct comparison, we also directly copy the first two rows from their original paper. We list four rows picked from the top-30 artists (Figure \ref{fig:top-30-artists}). The painting skills and styles of the artists are reflected. For example, in our first row all "drawn" by Vincent van Gogh, it is relatively easy to distinguish them from other artists: star sky appears often and the Zombie painting is telling a rich story of the author himself.

When the "street sign" is given in the first column, the original paper's two results mainly focused on the photo-style signs themselves. Yet, for the artists, the background nice street views are also important parts of the final painting, such as the sky, the forest, the building and the people with orange umbrella. With these hints, we modestly draw a preliminary conclusion that our four paints (rows 3 to 6) of the first column are more creative and includes richer sounding environment and humane information.

Column three, five and six are drawn from prompts which include ``fake objects'' which do not frequently exist in real-world. The ``half mouse half octopus'' is more like photos in the original paper (column 3, first 2 rows), our images are closer to hand-drawn paints. When drawing a ``chair that looks like an octopus'', all the rows in column six are close to artworks. 

The final column can be regarded as an industrial design oriented prompt. With the artists' style and genre included, we can positively imagine that when these paints are printed in real-world T-shirts, people will show their interests of further personalized customization and buy them.

\subsection{Textual Condition Extension Results}

\begin{figure*}
  \centering
  \includegraphics[width=16cm]{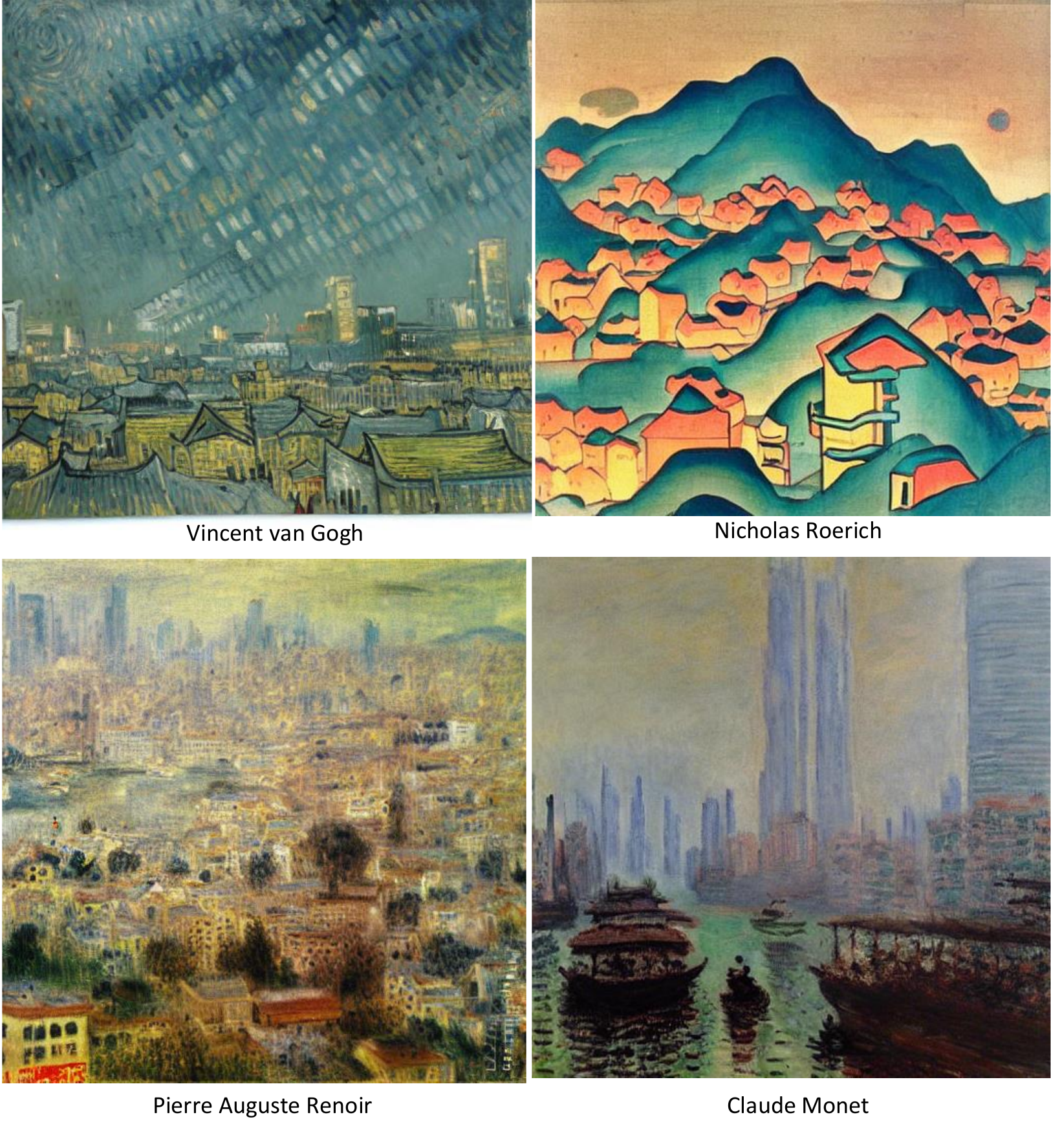}
  \caption{Four artists' artworks for the same prompt of ``a painting of urbanization of china''.}
  \label{fig:china_urban_4artists}
\end{figure*}

We use the former example of ``urbanization of China'' to show the results of textual condition extension. Figure \ref{fig:china_urban_4artists} shows four artworks by four famous artists, Vincent van Gogh, Nicholas Roerich, Pierre Auguste Renoir and Claude Monet. Interestingly, the major elements frequently used by artists are also reflected here. For example, the star sky of Vincent van Gogh, the water and boats of Claude Monet. The major elements included in the four paints are also interesting, combinations of Chinese traditional buildings and skycrapers, combinations of individual houses and mountains, rather crowded endless buildings and blurry sky, and Chinese traditional building style boats with super high skycrapers around the rivers.

\begin{figure*}
  \centering
  \includegraphics[width=16cm]{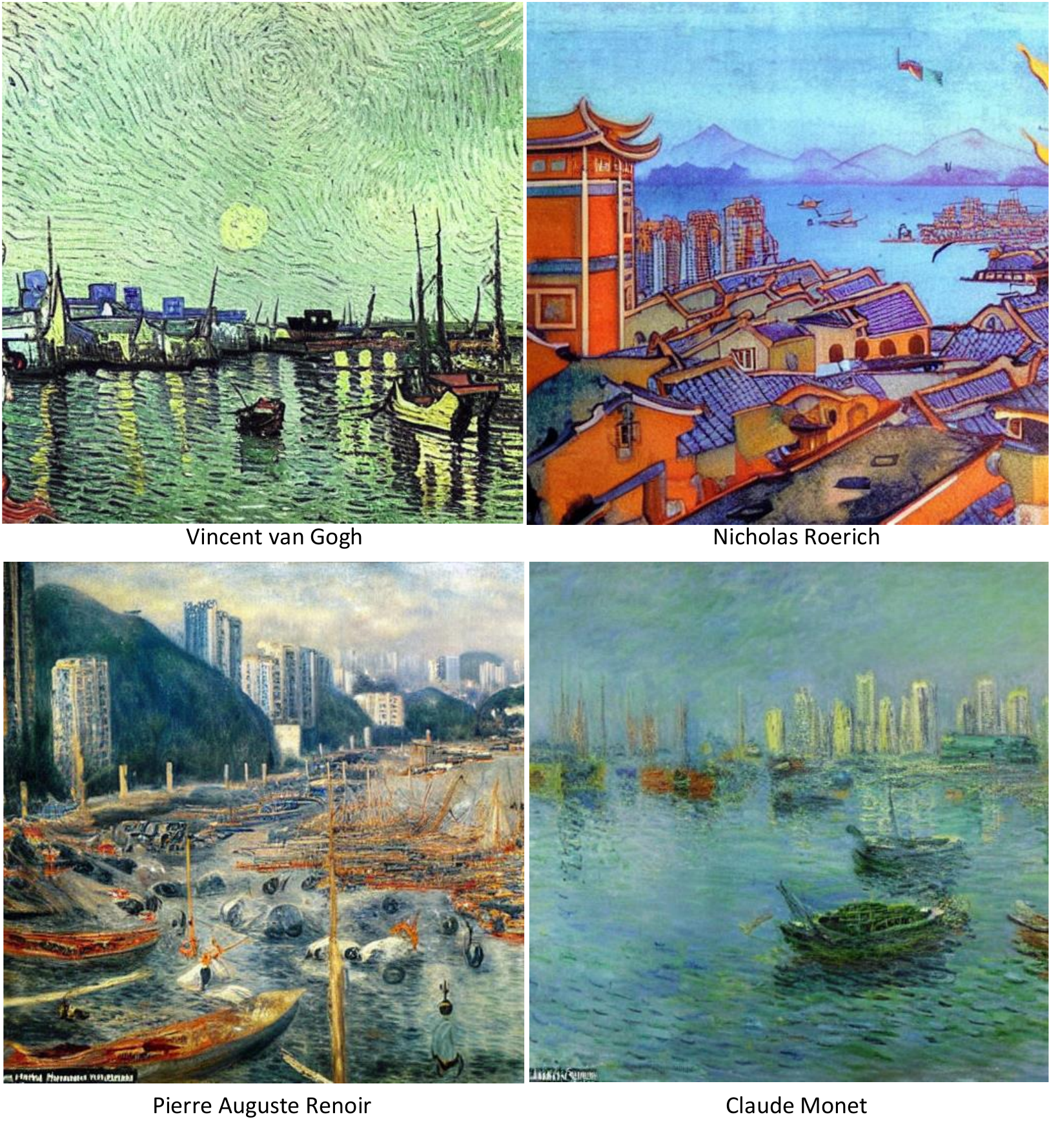}
  \caption{Four artists' artworks for the same extended prompt of ``originally a collection of fishing villages, Shenzhen rapidly grew to be one of the largest cities in China''.}
  \label{fig:china_urban_4artists_shenzhen1}
\end{figure*}

Figure \ref{fig:china_urban_4artists_shenzhen1} shows the same four artists' artwork for an extended prompt related to one of the most rapidly developed city, Shenzhen, during the urbanization of China. With the extended prompts, the model could generate more expressive images. For Vincent van Gogh, a moon in the middle of the sky, with fish ships near and high buildings in the far view. The same elments of fish boats and skycrapers are all included in the other three paintings. Interestingly, for Nicholas Roerich, even the skycrapers are drawn by following traditional Chinese style.

\begin{figure*}
  \centering
  \includegraphics[width=16cm]{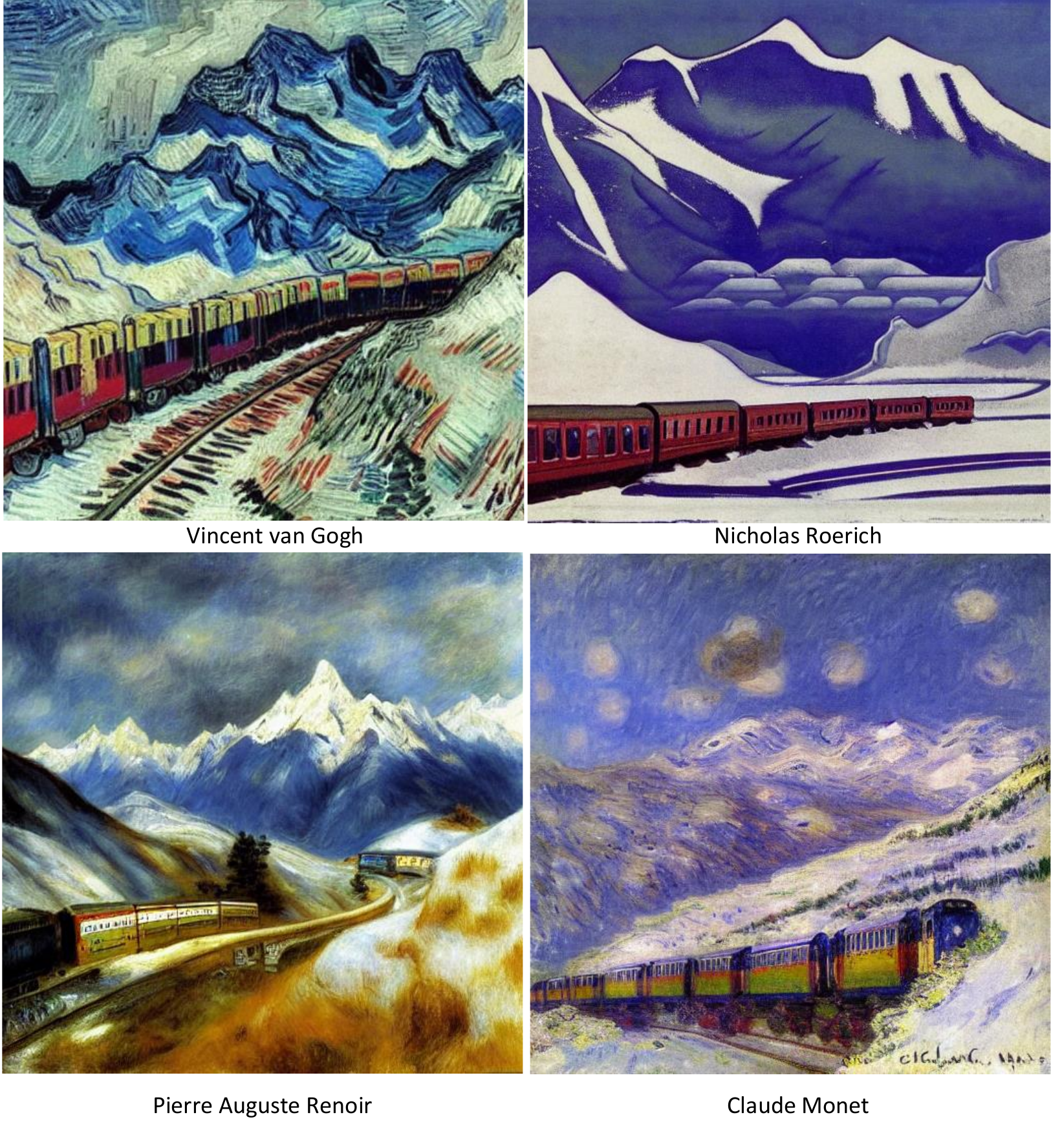}
  \caption{Four artists' artworks for the same extended prompt of ``a train runs on the snow-capped mountains of the Qinghai-Tibet Plateau''.}
  \label{fig:china_urban_4artists_train2}
\end{figure*}

Figure \ref{fig:china_urban_4artists_train2} shows the same four artists' artwork for an extended prompt related to a train running on the snow-capped mountains, during the urbanization of China. With the extended prompts, again, the model could generate more expressive images and keep the characters of each artists. The general styles and viewpoints of the four artists are reflected: now we have the mountain as the ``sky'' of Vincent van Gogh and the ``sky and mountain'' in Claude Monet looks like a reversed river.

\begin{figure*}
  \centering
  \includegraphics[width=16cm]{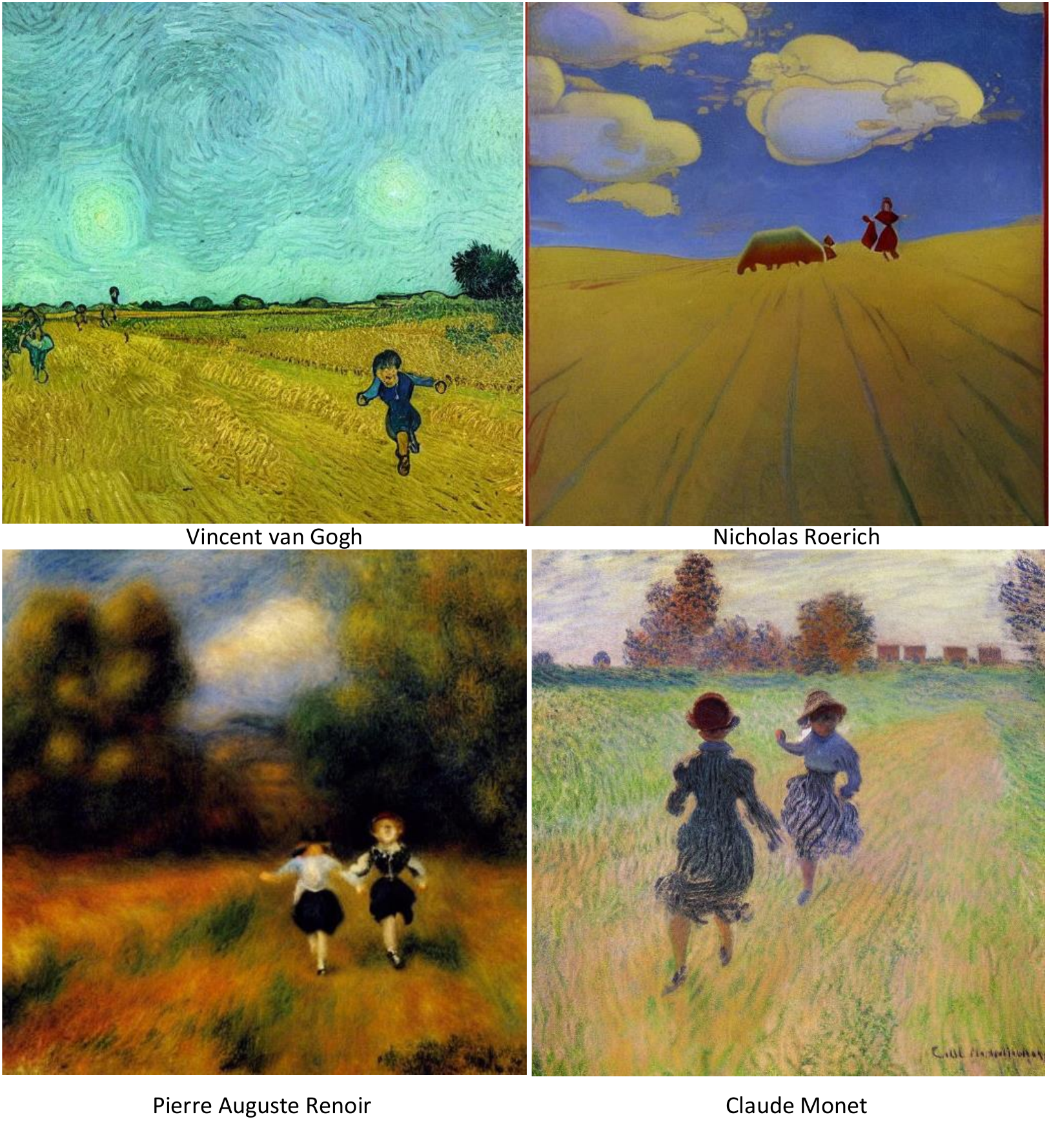}
  \caption{Four artists' artworks for the same extended prompt of ``left-behind children running in wheat-field''.}
  \label{fig:china_urban_4artists_children3}
\end{figure*}

Figure \ref{fig:china_urban_4artists_children3} shows the same four artists' artwork for an extended prompt related to children running in wheat-fields, during the urbanization of China. With the extended prompts, again, the model could generate more expressive images with rich emotional colors such as blue skys, golden wheat fields, and running-enjoy children. The general styles and viewpoints of the four artists are reflected, such as the Vincent van Gogh's sky, the skirts of the two girls from Claude Monet.

\begin{figure*}
  \centering
  \includegraphics[width=16cm]{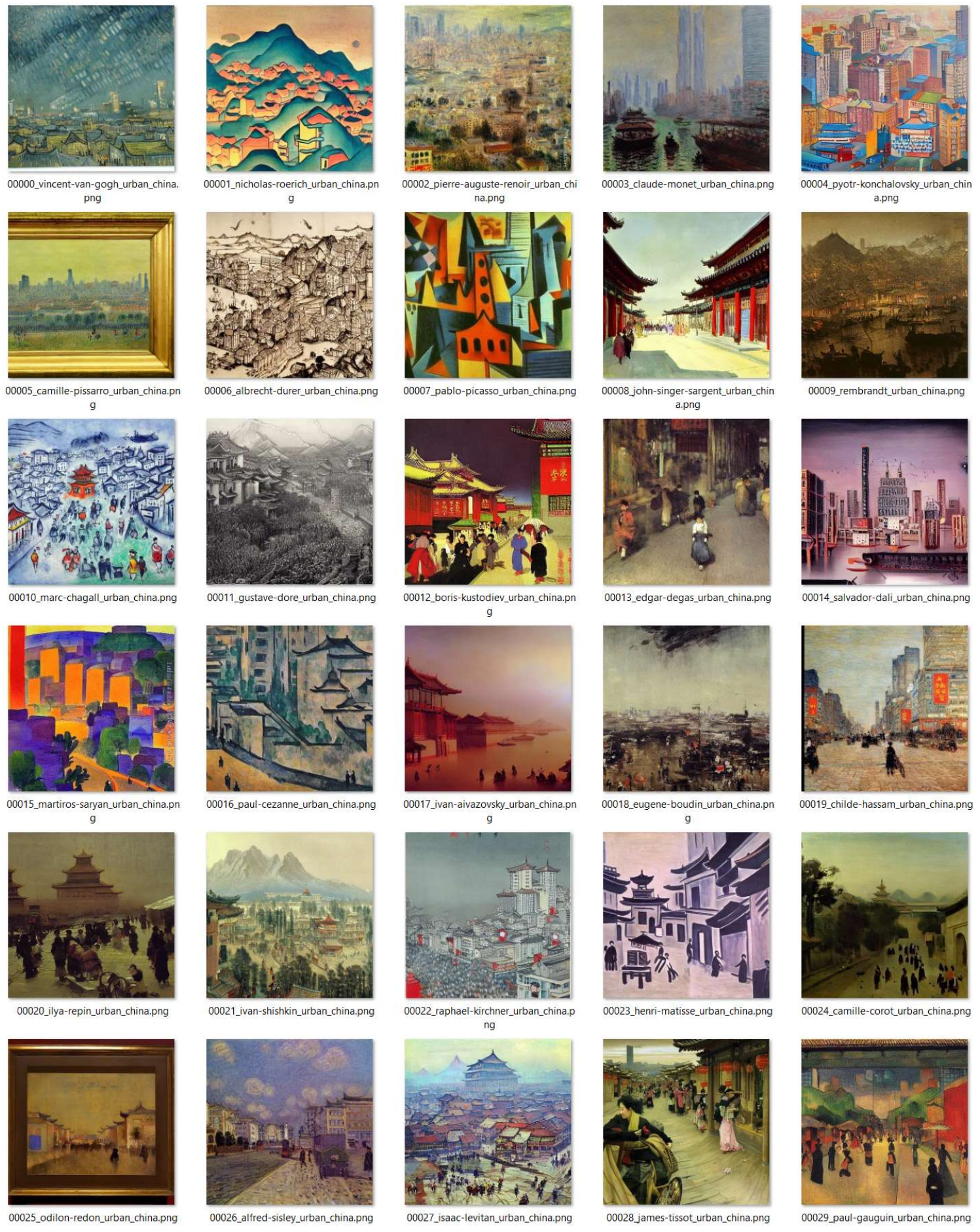}
  \caption{Top-30 artists' artworks for the same extended prompt of ``a painting of urbanization of china''.}
  \label{fig:china_urban_30artists_urban_china}
\end{figure*}

\begin{figure*}
  \centering
  \includegraphics[width=16cm]{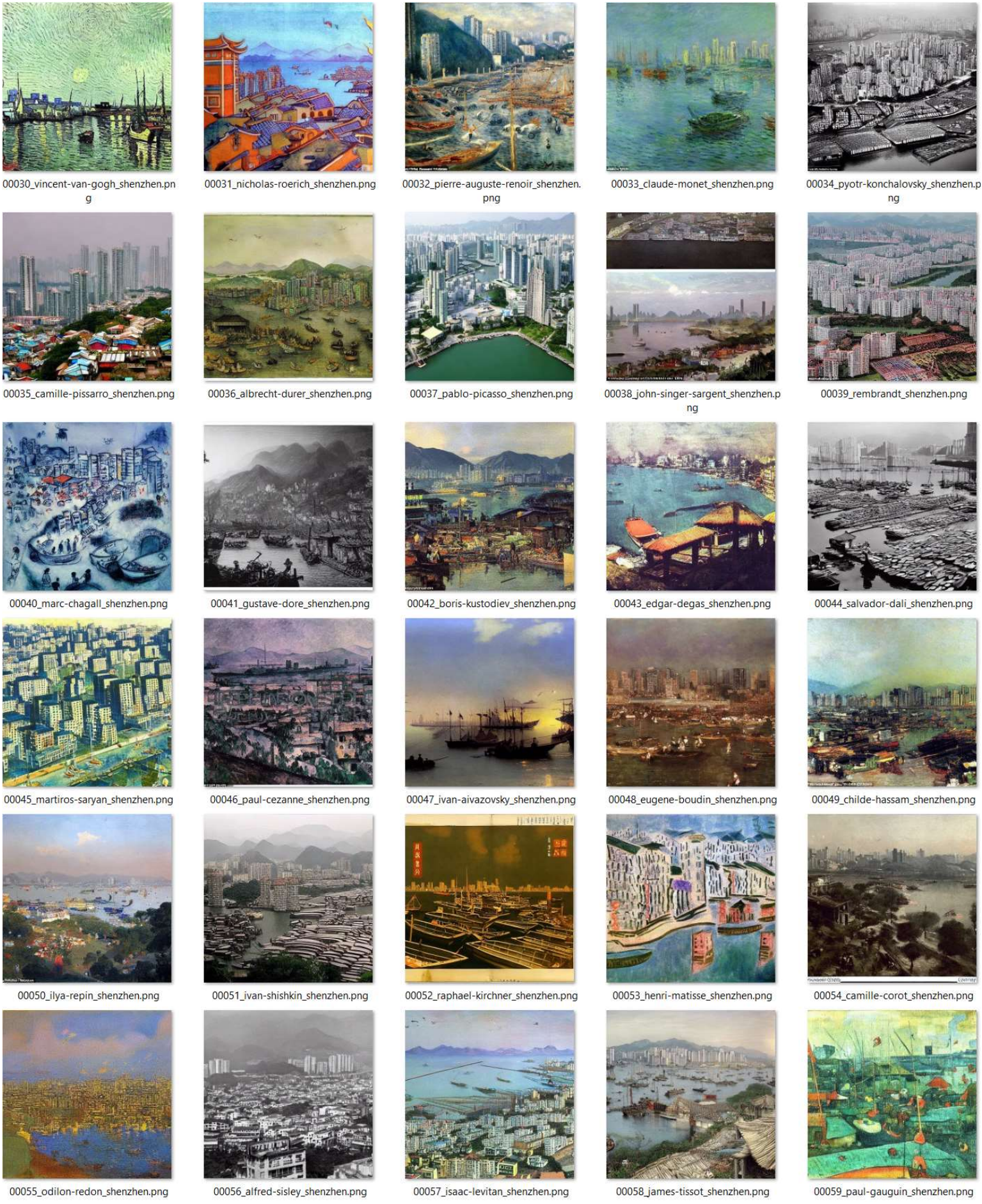}
  \caption{Top-30 artists' artworks for the same extended prompt of ``originally a collection of fishing villages, Shenzhen rapidly grew to be one of the largest cities in China''.}
  \label{fig:china_urban_30artists_shenzhen}
\end{figure*}

\begin{figure*}
  \centering
  \includegraphics[width=16cm]{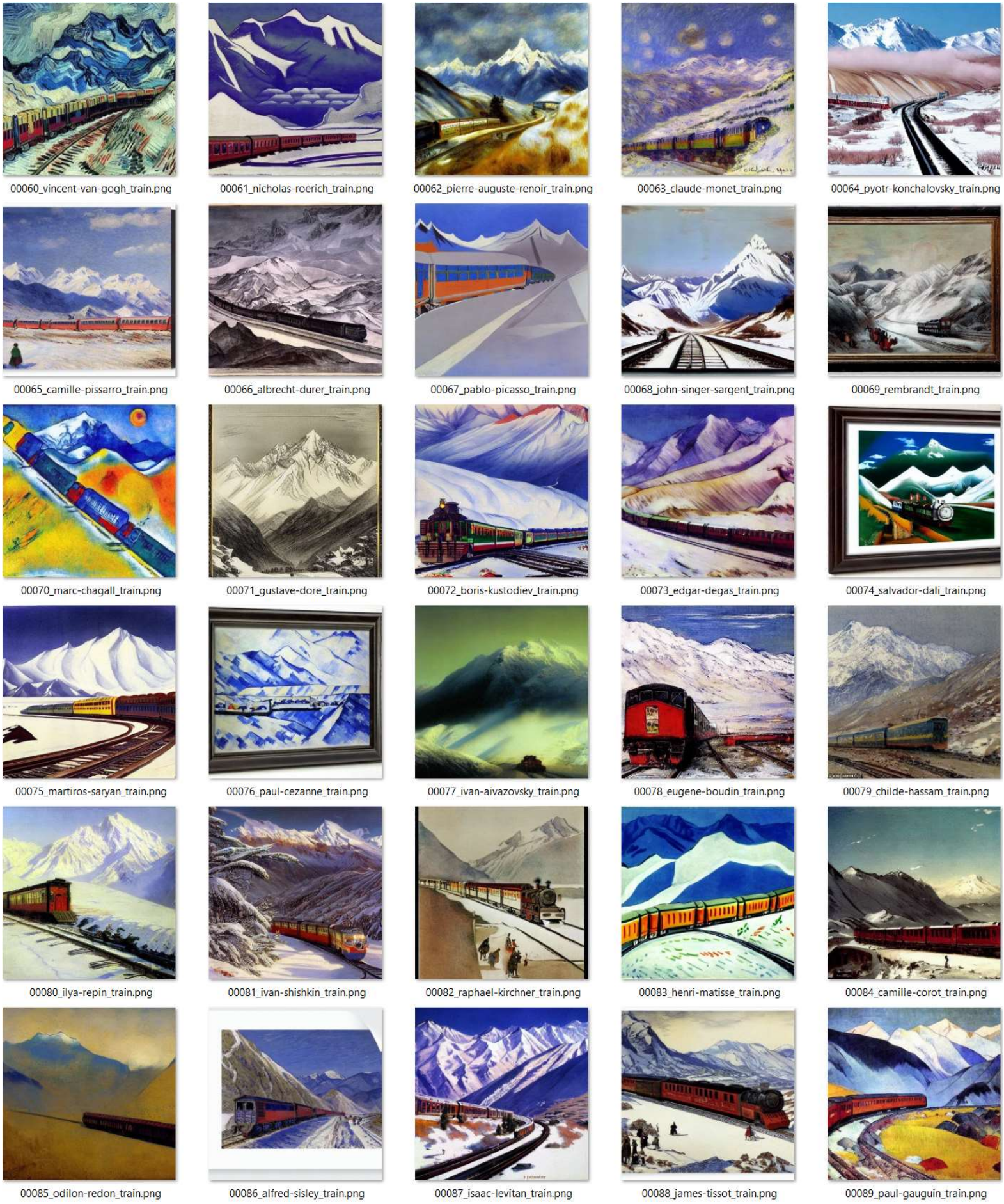}
  \caption{Top-30 artists' artworks for the same extended prompt of ``a train runs on the snow-capped mountains of the Qinghai-Tibet Plateau''.}
  \label{fig:china_urban_30artists_train}
\end{figure*}

\begin{figure*}
  \centering
  \includegraphics[width=16cm]{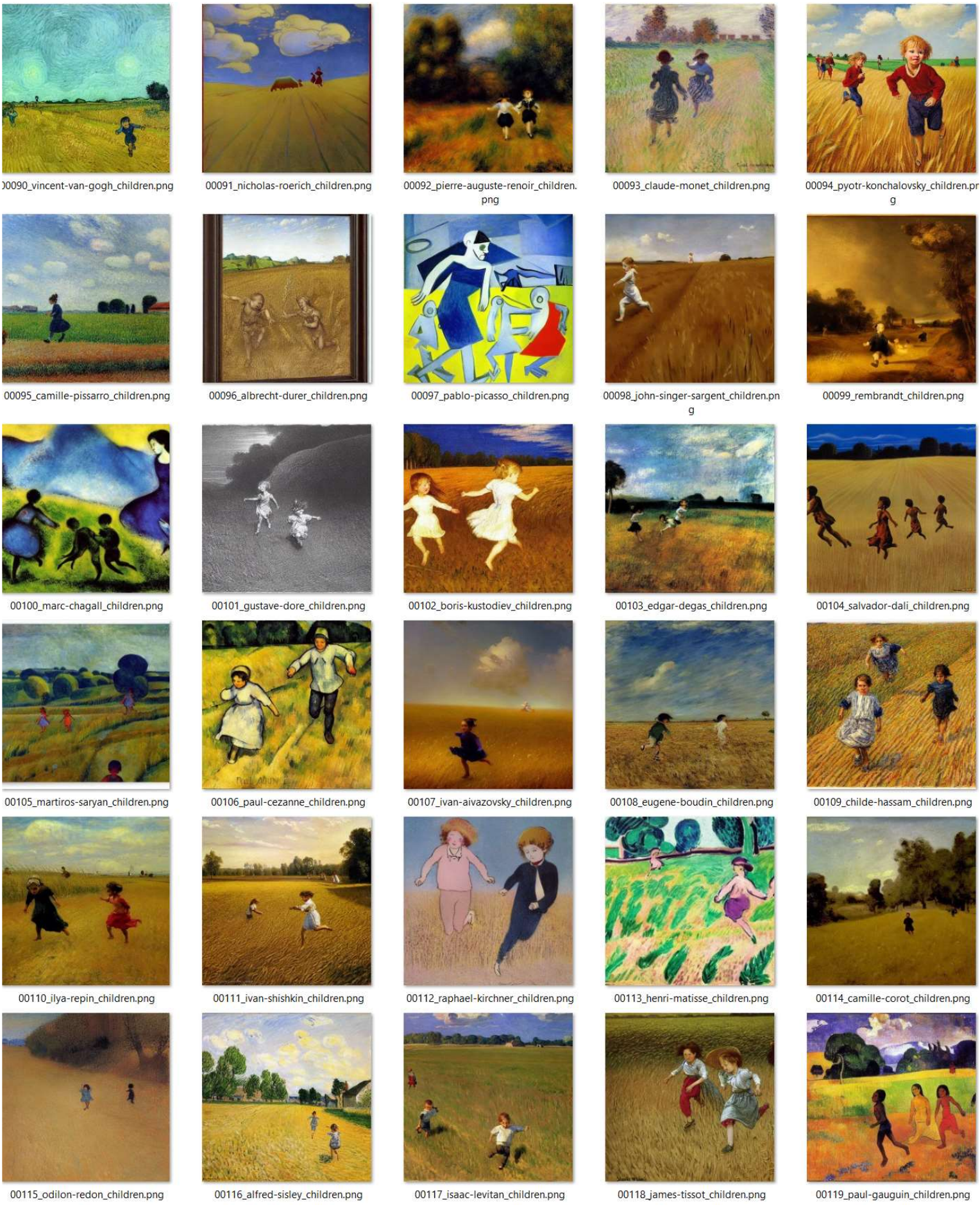}
  \caption{Top-30 artists' artworks for the same extended prompt of ``left-behind children running in wheat-field''.}
  \label{fig:china_urban_30artists_children}
\end{figure*}

Full images of the top-30 artists (Figure \ref{fig:compare-top-30-artists}) of the one initial prompt and three extended prompts are shown in Figure \ref{fig:china_urban_30artists_urban_china}, \ref{fig:china_urban_30artists_shenzhen}, \ref{fig:china_urban_30artists_train} and \ref{fig:china_urban_30artists_children} respectively.

\section{Conclusion}\label{sec:conclusion}

In order to improve the creativity of LDMs, we have proposed two directions of extending the input prompts and of retraining the original model by the Wikiart dataset. We take the 1,000 artists in recent 400 years as the major source of both creativity and artistry. With these proposals, the result diffusion models can ask these famous artists to draw novel and expressive paints of modern topics. 

We believe this is an interesting topic and has industrial design requirements for real-world applications, such as cloth designing, advertisement posters, and game character designing. Through drawing the real-world's topics with the help of hundreds to thousands famous artists, it is reasonable to learn the creativity and fertility from these artists' eyes.

\bibliography{anthology,custom}

\begin{thebibliography}{25}
\expandafter\ifx\csname natexlab\endcsname\relax\def\natexlab#1{#1}\fi

\bibitem[{Cao et~al.(2022)Cao, Tan, Gao, Chen, Heng, and
  Li}]{survey_generative_diffusion_https://doi.org/10.48550/arxiv.2209.02646}
Hanqun Cao, Cheng Tan, Zhangyang Gao, Guangyong Chen, Pheng-Ann Heng, and
  Stan~Z. Li. 2022.
\newblock \href {https://doi.org/10.48550/ARXIV.2209.02646} {A survey on
  generative diffusion model}.

\bibitem[{Croitoru et~al.(2022)Croitoru, Hondru, Ionescu, and
  Shah}]{survey_vision_diffusion_https://doi.org/10.48550/arxiv.2209.04747}
Florinel-Alin Croitoru, Vlad Hondru, Radu~Tudor Ionescu, and Mubarak Shah.
  2022.
\newblock \href {https://doi.org/10.48550/ARXIV.2209.04747} {Diffusion models
  in vision: A survey}.

\bibitem[{Esser et~al.(2020)Esser, Rombach, and
  Ommer}]{taming_DBLP:journals/corr/abs-2012-09841}
Patrick Esser, Robin Rombach, and Bj{\"{o}}rn Ommer. 2020.
\newblock \href {http://arxiv.org/abs/2012.09841} {Taming transformers for
  high-resolution image synthesis}.
\newblock \emph{CoRR}, abs/2012.09841.

\bibitem[{Ho et~al.(2020)Ho, Jain, and
  Abbeel}]{DDPM_DBLP:journals/corr/abs-2006-11239}
Jonathan Ho, Ajay Jain, and Pieter Abbeel. 2020.
\newblock \href {http://arxiv.org/abs/2006.11239} {Denoising diffusion
  probabilistic models}.
\newblock \emph{CoRR}, abs/2006.11239.

\bibitem[{Jabbar et~al.(2020)Jabbar, Li, and
  Omar}]{gan_survey1_DBLP:journals/corr/abs-2006-05132}
Abdul Jabbar, Xi~Li, and Bourahla Omar. 2020.
\newblock \href {http://arxiv.org/abs/2006.05132} {A survey on generative
  adversarial networks: Variants, applications, and training}.
\newblock \emph{CoRR}, abs/2006.05132.

\bibitem[{Jeong et~al.(2021)Jeong, Kim, Cheon, Choi, and
  Kim}]{diff_tts_https://doi.org/10.48550/arxiv.2104.01409}
Myeonghun Jeong, Hyeongju Kim, Sung~Jun Cheon, Byoung~Jin Choi, and Nam~Soo
  Kim. 2021.
\newblock \href {https://doi.org/10.48550/ARXIV.2104.01409} {Diff-tts: A
  denoising diffusion model for text-to-speech}.

\bibitem[{Liu et~al.(2022)Liu, Ren, Lin, and
  Zhao}]{pseudo_https://doi.org/10.48550/arxiv.2202.09778}
Luping Liu, Yi~Ren, Zhijie Lin, and Zhou Zhao. 2022.
\newblock \href {https://doi.org/10.48550/ARXIV.2202.09778} {Pseudo numerical
  methods for diffusion models on manifolds}.

\bibitem[{Liu et~al.(2021)Liu, Cao, Su, and
  Meng}]{sing_conv_diff_https://doi.org/10.48550/arxiv.2105.13871}
Songxiang Liu, Yuewen Cao, Dan Su, and Helen Meng. 2021.
\newblock \href {https://doi.org/10.48550/ARXIV.2105.13871} {Diffsvc: A
  diffusion probabilistic model for singing voice conversion}.

\bibitem[{Mittal et~al.(2021)Mittal, Engel, Hawthorne, and
  Simon}]{music_diffusion_DBLP:journals/corr/abs-2103-16091}
Gautam Mittal, Jesse~H. Engel, Curtis Hawthorne, and Ian Simon. 2021.
\newblock \href {http://arxiv.org/abs/2103.16091} {Symbolic music generation
  with diffusion models}.
\newblock \emph{CoRR}, abs/2103.16091.

\bibitem[{Popov et~al.(2021)Popov, Vovk, Gogoryan, Sadekova, and
  Kudinov}]{grad_tts_DBLP:journals/corr/abs-2105-06337}
Vadim Popov, Ivan Vovk, Vladimir Gogoryan, Tasnima Sadekova, and Mikhail~A.
  Kudinov. 2021.
\newblock \href {http://arxiv.org/abs/2105.06337} {Grad-tts: {A} diffusion
  probabilistic model for text-to-speech}.
\newblock \emph{CoRR}, abs/2105.06337.

\bibitem[{Radford et~al.(2021)Radford, Kim, Hallacy, Ramesh, Goh, Agarwal,
  Sastry, Askell, Mishkin, Clark, Krueger, and
  Sutskever}]{clip_DBLP:journals/corr/abs-2103-00020}
Alec Radford, Jong~Wook Kim, Chris Hallacy, Aditya Ramesh, Gabriel Goh,
  Sandhini Agarwal, Girish Sastry, Amanda Askell, Pamela Mishkin, Jack Clark,
  Gretchen Krueger, and Ilya Sutskever. 2021.
\newblock \href {http://arxiv.org/abs/2103.00020} {Learning transferable visual
  models from natural language supervision}.
\newblock \emph{CoRR}, abs/2103.00020.

\bibitem[{Raffel et~al.(2019)Raffel, Shazeer, Roberts, Lee, Narang, Matena,
  Zhou, Li, and Liu}]{t5_DBLP:journals/corr/abs-1910-10683}
Colin Raffel, Noam Shazeer, Adam Roberts, Katherine Lee, Sharan Narang, Michael
  Matena, Yanqi Zhou, Wei Li, and Peter~J. Liu. 2019.
\newblock \href {http://arxiv.org/abs/1910.10683} {Exploring the limits of
  transfer learning with a unified text-to-text transformer}.
\newblock \emph{CoRR}, abs/1910.10683.

\bibitem[{Robertson(2009)}]{bm25_robertson2009probabilistic}
S.~Robertson. 2009.
\newblock \href
  {http://scholar.google.de/scholar.bib?q=info:U4l9kCVIssAJ:scholar.google.com/&output=citation&hl=de&as_sdt=2000&as_vis=1&ct=citation&cd=1}
  {{The Probabilistic Relevance Framework: BM25 and Beyond}}.
\newblock \emph{Foundations and Trends{\textregistered} in Information
  Retrieval}, 3(4):333--389.

\bibitem[{Rombach et~al.(2021)Rombach, Blattmann, Lorenz, Esser, and
  Ommer}]{stablediffusion_DBLP:journals/corr/abs-2112-10752}
Robin Rombach, Andreas Blattmann, Dominik Lorenz, Patrick Esser, and
  Bj{\"{o}}rn Ommer. 2021.
\newblock \href {http://arxiv.org/abs/2112.10752} {High-resolution image
  synthesis with latent diffusion models}.
\newblock \emph{CoRR}, abs/2112.10752.

\bibitem[{Ronneberger et~al.(2015)Ronneberger, Fischer, and
  Brox}]{Unet_DBLP:journals/corr/RonnebergerFB15}
Olaf Ronneberger, Philipp Fischer, and Thomas Brox. 2015.
\newblock \href {http://arxiv.org/abs/1505.04597} {U-net: Convolutional
  networks for biomedical image segmentation}.
\newblock \emph{CoRR}, abs/1505.04597.

\bibitem[{Sohl{-}Dickstein et~al.(2015)Sohl{-}Dickstein, Weiss,
  Maheswaranathan, and Ganguli}]{DPM2015_DBLP:journals/corr/Sohl-DicksteinW15}
Jascha Sohl{-}Dickstein, Eric~A. Weiss, Niru Maheswaranathan, and Surya
  Ganguli. 2015.
\newblock \href {http://arxiv.org/abs/1503.03585} {Deep unsupervised learning
  using nonequilibrium thermodynamics}.
\newblock \emph{CoRR}, abs/1503.03585.

\bibitem[{Song et~al.(2020)Song, Meng, and
  Ermon}]{ddim_DBLP:journals/corr/abs-2010-02502}
Jiaming Song, Chenlin Meng, and Stefano Ermon. 2020.
\newblock \href {http://arxiv.org/abs/2010.02502} {Denoising diffusion implicit
  models}.
\newblock \emph{CoRR}, abs/2010.02502.

\bibitem[{Song and Ermon(2019)}]{ScoreMatching_NEURIPS2019_3001ef25}
Yang Song and Stefano Ermon. 2019.
\newblock \href
  {https://proceedings.neurips.cc/paper/2019/file/3001ef257407d5a371a96dcd947c7d93-Paper.pdf}
  {Generative modeling by estimating gradients of the data distribution}.
\newblock In \emph{Advances in Neural Information Processing Systems},
  volume~32. Curran Associates, Inc.

\bibitem[{Tan et~al.(2017)Tan, Chan, Aguirre, and
  Tanaka}]{artgan_DBLP:journals/corr/TanCAT17}
Wei~Ren Tan, Chee~Seng Chan, Hern{\'{a}}n~E. Aguirre, and Kiyoshi Tanaka. 2017.
\newblock \href {http://arxiv.org/abs/1702.03410} {Artgan: Artwork synthesis
  with conditional categorial gans}.
\newblock \emph{CoRR}, abs/1702.03410.

\bibitem[{Vaswani et~al.(2017)Vaswani, Shazeer, Parmar, Uszkoreit, Jones,
  Gomez, Kaiser, and Polosukhin}]{transformer_NIPS2017_3f5ee243}
Ashish Vaswani, Noam Shazeer, Niki Parmar, Jakob Uszkoreit, Llion Jones,
  Aidan~N Gomez, \L~ukasz Kaiser, and Illia Polosukhin. 2017.
\newblock \href
  {https://proceedings.neurips.cc/paper/2017/file/3f5ee243547dee91fbd053c1c4a845aa-Paper.pdf}
  {Attention is all you need}.
\newblock In \emph{Advances in Neural Information Processing Systems},
  volume~30. Curran Associates, Inc.

\bibitem[{Wang et~al.(2021)Wang, She, and Ward}]{gan_survey2}
Zhengwei Wang, Qi~She, and Tomas Ward. 2021.
\newblock \href {https://doi.org/10.1145/3439723} {Generative adversarial
  networks in computer vision: A survey and taxonomy}.
\newblock \emph{ACM Computing Surveys}, 54:1--38.

\bibitem[{Wolleb et~al.(2022)Wolleb, Bieder, Sandkühler, and
  Cattin}]{medical_diff_https://doi.org/10.48550/arxiv.2203.04306}
Julia Wolleb, Florentin Bieder, Robin Sandkühler, and Philippe~C. Cattin.
  2022.
\newblock \href {https://doi.org/10.48550/ARXIV.2203.04306} {Diffusion models
  for medical anomaly detection}.

\bibitem[{Xue et~al.(2022)Xue, Wang, Zhang, Xie, Zhu, and
  Bi}]{learn2sing_https://doi.org/10.48550/arxiv.2203.16408}
Heyang Xue, Xinsheng Wang, Yongmao Zhang, Lei Xie, Pengcheng Zhu, and Mengxiao
  Bi. 2022.
\newblock \href {https://doi.org/10.48550/ARXIV.2203.16408} {Learn2sing 2.0:
  Diffusion and mutual information-based target speaker svs by learning from
  singing teacher}.

\bibitem[{Yang et~al.(2022)Yang, Zhang, Hong, Xu, Zhao, Shao, Zhang, Yang, and
  Cui}]{diff_beida_survey_https://doi.org/10.48550/arxiv.2209.00796}
Ling Yang, Zhilong Zhang, Shenda Hong, Runsheng Xu, Yue Zhao, Yingxia Shao,
  Wentao Zhang, Ming-Hsuan Yang, and Bin Cui. 2022.
\newblock \href {https://doi.org/10.48550/ARXIV.2209.00796} {Diffusion models:
  A comprehensive survey of methods and applications}.

\bibitem[{Zhang et~al.(2019)Zhang, Sun, Galley, Chen, Brockett, Gao, Gao, Liu,
  and Dolan}]{dialogpt_DBLP:journals/corr/abs-1911-00536}
Yizhe Zhang, Siqi Sun, Michel Galley, Yen{-}Chun Chen, Chris Brockett, Xiang
  Gao, Jianfeng Gao, Jingjing Liu, and Bill Dolan. 2019.
\newblock \href {http://arxiv.org/abs/1911.00536} {Dialogpt: Large-scale
  generative pre-training for conversational response generation}.
\newblock \emph{CoRR}, abs/1911.00536.

\end{thebibliography}
\bibliographystyle{acl_natbib}




\end{document}